\documentclass[sigplan,10pt,screen]{acmart}
\renewcommand\footnotetextcopyrightpermission[1]{}
\usepackage{xspace}
\usepackage{tabularx}
\usepackage{subcaption}
\usepackage{enumitem}
\usepackage{algorithm}
\usepackage{algpseudocode}
\usepackage{pifont}
\usepackage{multirow}
\usepackage{booktabs}
\usepackage{graphicx}
\usepackage{etoolbox}
\usepackage[all]{nowidow}
\usepackage{colortbl}
\usepackage{latexsym}
\usepackage{svg}
\usepackage{rotating}
\usepackage{amsmath}

\setitemize{noitemsep,topsep=0pt,parsep=0pt,partopsep=0pt}

\newcommand{\parabf}[1]{\medskip\noindent\textbf{#1}}
\newcommand{\parait}[1]{\medskip\noindent\textit{#1}}
\newcommand{\paraf}[1]{\noindent\textbf{#1}}
\newcommand{\cut}[1]{}

\newcommand{\sysname}{StreamRL\xspace}

\begin{document}
\title{\sysname: Scalable, Heterogeneous, and Elastic 
        RL for LLMs with Disaggregated Stream Generation}
\pagestyle{plain}
\author{
    \rm{
        Yinmin Zhong$^{\text{1}}$ \enskip
        Zili Zhang$^{\text{1}}$ \enskip
        Xiaoniu Song$^{\text{2}}$ \enskip
        Hanpeng Hu$^{\text{2}}$ \enskip
    }
    \\
    \rm{
        Chao Jin$^{\text{1}}$ \enskip
        Bingyang Wu$^{\text{1}}$ \enskip
        Nuo Chen$^{\text{2}}$ \enskip
        Yukun Chen$^{\text{2}}$ \enskip
        Yu Zhou$^{\text{2}}$ \enskip
    }
    \\
    \rm{
        Changyi Wan$^{\text{2}}$ \enskip
        Hongyu Zhou$^{\text{2}}$ \enskip
        Yimin Jiang$^{\text{3}}$ \enskip
        Yibo Zhu$^{\text{2}}$ \enskip
        Daxin Jiang$^{\text{2}}$ \enskip
    }
    \\
    \vspace{0.2in}
    {$^{\text{1}}$\textit{School of Computer Science, Peking University}\enskip $^{\text{2}}$\textit{StepFun}\enskip $^{\text{3}}$\textit{Unaffiliated}}
}

\newcommand{\SGS}{\texttt{SGS}\xspace}
\newcommand{\Trainer}{\texttt{Trainer}\xspace}
\begin{abstract}
    Reinforcement learning (RL) has become the core post-training technique 
    for large language models (LLMs). 
    RL for LLMs involves two stages: generation and training. 
    The LLM first generates samples online, which are then used to derive rewards for training. 
    The conventional view holds that the \textit{colocated architecture}—where
    the two stages share resources via temporal multiplexing—outperforms the 
    \textit{disaggregated architecture}, in which dedicated resources are assigned to each stage. 
    However, in real-world deployments, we observe that the colocated architecture 
    suffers from \textit{resource coupling}, where the two stages are constrained to use the same resources. 
    This coupling compromises the scalability and cost-efficiency of colocated RL in large scale training.
    In contrast, the disaggregated architecture allows for flexible resource allocation, 
    supports heterogeneous training setups, and facilitates cross-datacenter deployment.

    \sysname is designed with disaggregation from first principles and fully 
    unlocks its potential by addressing two types of performance bottlenecks 
    in existing disaggregated RL frameworks: 
    \textit{pipeline bubbles}, caused by stage dependencies, and 
    \textit{skewness bubbles}, resulting from long-tail output length distributions.
    To address pipeline bubbles, \sysname breaks the traditional stage boundary 
    in synchronous RL algorithms through stream generation, 
    and achieves fully overlapping in asynchronous RL.
    To address skewness bubbles, \sysname employs an output-length ranker model 
    to identify long-tail samples and reduces generation time 
    via skewness-aware dispatching and scheduling.
    Experiments show that \sysname improves throughput by up to $2.66\times$ 
    compared to existing state-of-the-art systems, and improves cost-effectiveness 
    by up to $1.33\times$ in heterogeneous, cross-datacenter setting.
\end{abstract}

\maketitle

\section{Introduction}
\label{sec:introduction}

Reinforcement learning (RL) has emerged as a new paradigm for training large language models (LLMs), 
substantially improving their reasoning capabilities and revealing a novel inference-time scaling law~\cite{deepseekr1, yu2025dapo, seed-thinking}.
State-of-the-art models such as OpenAI o1~\cite{openaio1} and o3~\cite{openaio3}, Claude 3.7 Sonnet~\cite{claude37}, 
and DeepSeek-R1~\cite{deepseekr1} have all adopted RL to achieve leading performance in tasks such as 
coding and mathematics.

In contrast to traditional next-token prediction~\cite{vaswani2017attention, brown2020language, openai2023gpt4, chowdhery2022palm, touvron2023llama} in pre-training, RL enables the 
LLMs to learn by trial and error from reward signals.
While numerous RL algorithms exist, such as PPO~\cite{schulman2017proximal} and GRPO~\cite{shao2024deepseekmath}, 
the typical RL workflow for LLMs involves two main stages in serial: generation and training.
In generation stage, the LLM produces samples on a batch of given prompts,
followed by the training stage that updates the LLM based on rewards 
derived from the generated samples. 

\begin{figure}[t!]
    \centering
    \includegraphics[width=0.95\linewidth]{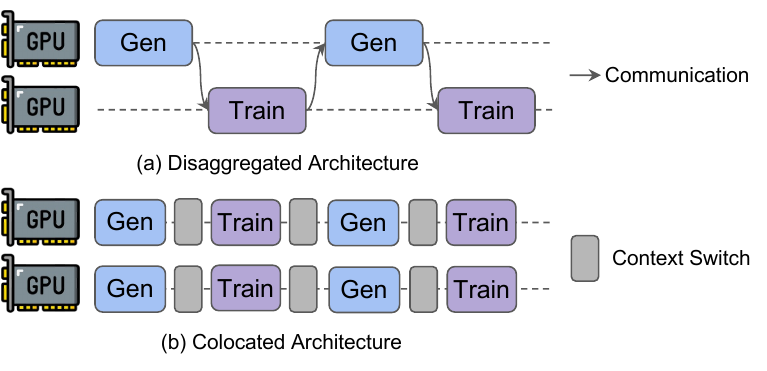}
    \vspace*{-0.1in}
    \caption{Two representative RL framework architectures.}
    \vspace*{-0.2in}
    \label{fig:intro:teaser}
\end{figure}

Due to this two-stage workflow, initial RL training frameworks for LLMs,
such as OpenRLHF~\cite{hu2024openrlhf} and NeMo~\cite{Harper_NeMo_a_toolkit},
naturally adopted a \textit{disaggregated architecture}.
As depicted in Figure~\ref{fig:intro:teaser}(a), dedicated computational resources are allocated
for each stage separately. The generation stage employs existing inference
frameworks like vLLM~\cite{vllm} to generate samples, which are subsequently transferred to the
training stage that uses training frameworks like DeepSpeed~\cite{aminabadi2022deepspeed} or Megatron-LM~\cite{shoeybi2020megatronlm}. 
Updated model weights are then transferred back to the inference framework for the
next iteration generation.
This architectural choice can effectively reuse existing infrastructures
and facilitate rapid deployment of various RL algorithms,
which gains broad initial adoption.
However, a notable drawback of disaggregated architecture is resource idleness
arising from serial dependency:
GPU resources allocated to the training stage remain idle during the generation stage,
and vice versa.

To address this inefficiency, recent RL training frameworks, such as verl~\cite{sheng2024hybridflow},
ReaL~\cite{mei2024realhf}, and RLHFuse~\cite{zhong2024rlhfuseefficientrlhftraining},
have adopted a \textit{colocated architecture}.
As illustrated in Figure~\ref{fig:intro:teaser}(b), it colocates the generation and training stages 
on the same GPU resources. When one stage 
is active, the states of the other stage (such as model weights and optimizer states)
is temporarily stored in CPU memory. Context switching happens between stage 
boundary, enabling time-division multiplexing of GPU resources.
This architectural shift resolves the resource idleness issue and improves training efficiency.
Subsequently, colocation becames the prevailing choice and was widely acknowledged to be superior 
to disaggregation.

We initially believed the same and chose colocated architecture for our 
internal RL framework. However, in practical deployment, we have observed that the 
colocated architecture encounters the problem of \textit{resource coupling} as the training scales out.
The reason lies in that the two stages feature fundamentally
distinct workloads: the generation stage is notably memory-bandwidth-bound~\cite{zhong2024distserve},
whereas the training stage is typically compute-bound~\cite{jiang2024megascale}.
However, due to colocation, both stages must share identical
\textit{resource quantities} and \textit{hardware types}, creating an inherent conflict with their divergent computational
characteristics.

Concretely, the performance speedup for the generation stage
reaches a plateau much more quickly compared to the compute-bound training stage
when scaling out resources.
Yet, the colocated architecture restricts resource quantities for both stages to be identical,
thus diminishing overall resource utilization. Also, it is unable to select 
the most suitable and cost-effective hardware types for each stage
respectively.
Moreover, constrained by factors such as policy, cost, and power supply, constructing 
a single large-scale datacenter can be challenging and expensive~\cite{multi_dc_report2}. 
Consequently, companies typically operate multiple medium-sized datacenters 
equipped with GPUs spanning different generations and types, forming a 
cross-datacenter heterogeneous resource pool~\cite{multi_dc_report1}. 
As RL training scales out, 
the colocated architecture struggles to efficiently leverage this entire resource pool, 
as the training stage typically involves full-mesh communication operations~\cite{infinitepod}, 
which will incur significant communication overhead across datacenters.

Under such circumstances, the initial disaggregated architecture reshines.
First, resource quantities for the generation and training stages need not 
be identical, allowing more flexible and efficient resource allocation. Second, 
it allows for selecting the most suitable hardware for each stage, such as
adopting more cost-effective inference GPUs for the generation stage,
while using more expensive and high-performance GPUs exclusively for the training stage.
Moreover, RL features point-to-point data transfer between the two stages,
keeping the communication overhead manageable even across datacenters.
By putting the two stages in separate datacenters, RL training 
can overcome the constraints of a single homogeneous datacenter and fully 
leverage the entire cross-datacenter, heterogeneous resource pool.

Nevertheless, the disaggregated architecture encounters two challenges 
in existing frameworks, preventing it from fully unleashing its potential.
The first is the resource idleness as illustrated in Figure~\ref{fig:intro:teaser}(a).
Naive disaggregation introduces pipeline bubbles due to the serialized execution of the two stages.  
The second challenge, irrespective of the underlying architecture, stems from the long-tail output 
length distribution inherent in LLM inference workload~\cite{zhong2024rlhfuseefficientrlhftraining}. 
In the later phase of generation, only a small number of long-tail samples remain in the system,
leading to severely under-utilized GPUs. This problem is further exacerbated by the widespread use 
of long chain-of-thought generation in modern RL training for reasoning models~\cite{deepseekr1}.  

To address the above problems, we present \sysname,
an RL training framework specifically designed for the disaggregated architecture. 
The key insight of \sysname is abstracting the generation and training stages into 
\textit{stream generation service} (\SGS) and \Trainer, respectively.
\Trainer submits generation requests to \SGS; once the \SGS receives prompts and starts generation, 
it returns each completed sample to \Trainer in a stream fashion, 
allowing any follow-up actions of \Trainer and reducing resource 
idleness.
With streaming, \sysname can enhance existing naive solutions such as mini-batch 
pipelining~\cite{deepcoder2025} to enable more flexible and efficient concurrent 
execution of \SGS and \Trainer, and can achieve fully overlapping execution 
under asynchronous RL.

Under this streaming architecture, careful resource allocation between the
two stages is necessary to balance their execution time, otherwise
pipeline bubbles may still occur. \sysname utilize a \textit{profiler-based
resource allocation algorithm} to decide the resource allocation before training.
Furthermore, some recent progress in reasoning LLMs~\cite{deepseekr1} has witnessed improvements in model capabilities during RL training 
as the LLM output lengths increase. Given that the workload change sensitivity 
of the two stages differs, \sysname also provides a \textit{dynamic resource adjustment
mechanism} to elastically maintain balanced execution between the two stages throughout
the training process.

To address the long-tail issue, \sysname leverages an \textit{output-length ranker model}
to identify long-tail samples, then utilizes a \textit{skewness-aware scheduling mechanism} 
that selectively allocates resources to long-tail samples and adjusts the batch size 
to reduce overall generation latency.

Experiments on various LLMs and real-world dataset show that \sysname achieves 
up to $2.66\times$ throughput compared to existing state-of-the-art systems, 
and can further improves up to $1.33\times$ in cost-effectiveness under heterogeneous, 
cross-datacenter setting.

In summary, we make the following contributions:
\begin{itemize}[leftmargin=*]
\item We analyze critical scalability and efficiency issues inherent
      in current colocated RL frameworks, and propose to 
      revisit the disaggregated architecture.
\item We present \sysname to effectively mitigate pipeline bubbles 
      and the long-tail issue to fully unleash the potential of disaggregated architecture.
\item We conduct extensive experiments to evaluate \sysname's performance against current 
      state-of-the-art RL frameworks and demonstrate its effectiveness in heterogeneous, 
      cross-datacenter scenarios.
\end{itemize}

\section{Background and Motivation}

\subsection{Background}
\label{sec:background}

\paraf{RL for LLMs.}
We first explain how key concepts in RL are applied to LLM training.
The LLM to be trained works as the \textit{agent model} in the RL
semantics, which learns from feedback from interactions with the environment.
The prompt represents the \textit{initial state} in which the agent model resides.
Each token generated by the agent model is treated as an \textit{action},
which leads to a new state—namely, the combination of the prompt and all
the generated tokens.
The agent model continues to take actions in each state, autoregressively
generating tokens until completing a sample, which is referred to as 
a \textit{trajectory} or \textit{rollout}.
Subsequently, the \textit{reward} is derived from the samples
and used to train the agent model.

Concretely, each RL iteration consists of two primary stages: generation and training.
In the generation stage, for each given prompt, the agent model first processes all the
tokens through one forward pass to generate the first output token, which is called 
the prefill phase. 
It also establishes the key-value cache~\cite{ott2019fairseq, dai2019transformer}, which stores per-token intermediate states.
Then, in the decoding phase, the model autoregressively generates subsequent tokens,
reusing the key-value cache to avoid redundant computations.
Upon completion, each prompt, combined with the generated tokens, forms a single
training sample. For sample efficiency, hundreds of samples are generated in batch 
for each iteration.

In the training stage, generated samples are scored with reward. Depending on the algorithm,
reward computation can be performed by an additionally trained \textit{Reward Model}~\cite{bai2022traininghelpfulharmlessassistant}
or by rule-based functions~\cite{deepseekr1}. The latter approach is commonly used for tasks with explicit correctness
criteria, such as coding and solving maths problems, and recent research has demonstrated its
effectiveness in improving model's reasoning ability~\cite{deepseekr1, seed-thinking, yu2025dapo}.
In addition to coarse-grained, sample-level reward, certain RL algorithms, 
such as PPO~\cite{schulman2017proximal}, introduce a \textit{Critic Model} to provide fine-grained, 
action-level reward for each token.
In the constrast, other algorithms like GRPO~\cite{shao2024deepseekmath} use approximation strategies to avoid using Critic Model.
To enhance training stability, a \textit{Reference Model}—initialized 
from the before-trained agent model and remained frozen 
during training—is used to provide Kullback-Leibler (KL) divergence regularization.
It prevents the agent model from deviating much during training. 
The final loss incorporates both sample- and token-level rewards 
along with KL divergence to train the agent model. 
After updating parameters, it proceeds to the next iteration's generation.

\parabf{LLM Parallelization.} To scale up LLM training, several parallelization 
techniques have been developed. Data Parallelism (DP) involves duplicating the model 
across devices and splitting the dataset among them, allowing each replica to process 
different portions of data concurrently. After each training step, gradients must be 
synchronized among all replicas. Tensor Parallelism (TP) divides individual operations 
across multiple GPUs, with each GPU responsible for a portion of the computation. 
Due to its high communication overhead, TP is typically confined to intra-node deployment 
where high-speed interconnects like NVLINK are available. Pipeline Parallelism (PP) 
splits the model layers into separate stages, assigning each to a different device or node.
The input batch is further divided into microbatches, enabling pipelined execution and 
full-batch gradient accumulation. Since each method offers distinct trade-offs, 
modern training systems~\cite{shoeybi2019megatron, jiang2024megascale} often combine all three approaches to 
maximize scalability and efficiency.

\subsection{Problems with Colocation}
\label{sec:background:problems}
As mentioned in \S\ref{sec:introduction}, the colocated architecture for RL framework,
with its advantage in resource efficiency, appears to be a better choice.
We also held this belief at the outset and
built our internal RL training framework based on colocated architecture to support the 
large-scale training of commercial models. 
However, with real-world deployment over time, 
as model size and training scale continues to grow, 
we found that the intuitively better colocated architecture began to reveal
its fundamental limitation, i.e., \textit{resource coupling}.

In the colocated architecture, the generation and training stages must share the same 
set of devices. However, the fundamental issue lies in the fact that these two stages 
represent fundamentally different workloads. 
During the decoding phase of the generation stage, each sample computes only on the 
newly generated token from the previous step, but still requires access to the 
full set of model parameters, making it highly memory-bandwidth-bound.
In contrast, the training stage is compute-bound,
as both forward and backward passes compute over all tokens in the batch 
simultaneously, easily saturating the GPU's compute units.
Due to colocation, both stages must share identical
\textit{resource quantities} and \textit{hardware types}, creating an inherent
conflict with their divergent computational characteristics.

\begin{figure}[t!]
    \centering
    \includegraphics[width=0.95\linewidth]{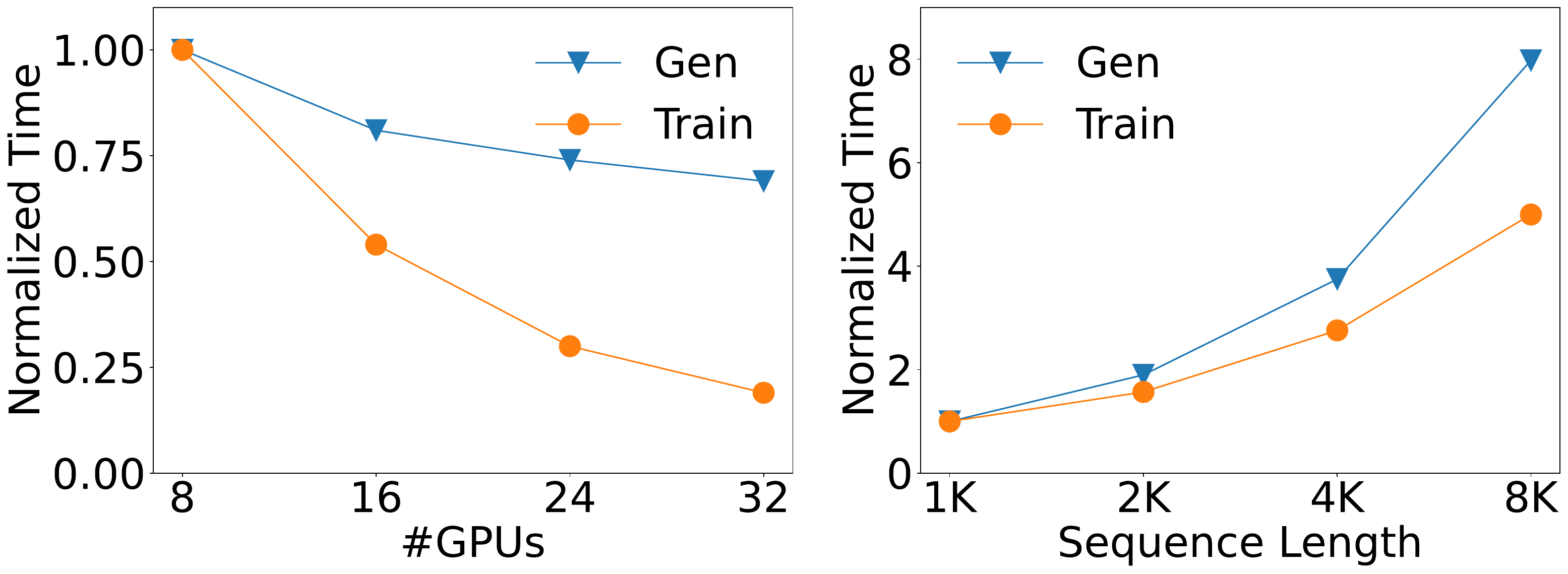}
    \vspace*{-0.1in}
    \caption{The performance sensitivity difference of the generation and training stage
    under resource quantities (left) and sequence length (right).}
    \vspace*{-0.2in}
    \label{fig:background:motivation}
\end{figure}

\parabf{Resource quantities.} Given the workload, we expect the iteration time to decrease with
more resources. However, due to the distinct computational characteristics
of the generation and training stages, their scaling sensitivity to resource quantities differ
significantly. We profile the execution latency of each stage under different resources 
when training a 7B LLM on a fixed workload with a sequence length of 8192.
As shown in the left of Figure~\ref{fig:background:motivation}, generation time quickly 
reaches a plateau as resources increase. This is because, being memory-bandwidth-bound,
the generation time is mainly determined by the overall memory bandwidth. 
Among parallelism strategies, only increasing 
the tensor parallelism (TP) size can effectively increase the overall bandwidth. 
However, due to the high communication overhead of TP, 
it is typically limited to intra-node where NVLINK are available.
As a result, scaling the resources for generation stage mainly increases
the number of generation instances, i.e., the DP size, which has limited effect on 
reducing generation time.

In contrast, since the training stage is compute-bound, it benefits much more from 
resource scaling, achieving better acceleration. The consequence of this difference 
in scaling sensitivity is that when scaling up resources, the generation stage suffers 
from low resource utilization, as increased resources do not translate into proportional 
performance gains. This issue becomes increasingly prominent as long chain-of-thought 
generation grows in importance, leading to longer model outputs and a rising share 
of generation time in the overall iteration time.

\parabf{Hardware types.} 
Another manifestation of resource coupling lies in hardware selection. 
As shown in Table~\ref{table:gpu_specs}, different NVIDIA GPU types exhibit 
trade-offs between compute capability, memory bandwidth and cost. Some GPUs, 
such as H20, are specifically designed for memory-bandwidth-bound workloads 
like inference, offering even higher HBM bandwidth and larger HBM capacity 
than flagship ones like H800, while costing only about 35\% as much. 
Therefore, from the perspective of training cost —e.g., 
throughput per cost—the colocated architecture prevents selecting the most 
cost-effectiveness hardware for each stage.

\subsection{Motivations for Disaggregation}
\label{sec:background:motivation}

In contrast, the initial disaggregated architecture exhibits several 
unique advantages that deserves reconsideration.

\parabf{Flexibility.}  
Under the disaggregated architecture, the resource coupling problem mentioned 
above is immediately eliminated, allowing dedicated resource allocation tailored 
to the distinct workloads of the two stages. Additionally, it enables selecting 
the most suitable hardware for each stage, flexibly leveraging heterogeneous 
resources to improve training cost and overall cost-effectiveness.

\begin{table}[t]
    \centering
    \begin{tabular}{lcc}
        \hline
            & H20 & H800 \\
        \hline
        BF16 TFLOPS       & 148        & \bf{989.5} \\
        HBM capacity      & \bf{96GB}  & 80GB       \\
        HBM bandwidth     & \bf{4TB/s} & 3.35TB/s   \\
        NVLINK bandwidth  & \bf{900GB/s} & 400GB/s \\
        Cost per machine~\cite{zhu2025megascaleinferservingmixtureofexpertsscale} & 1.00 & \bf{2.85} \\
        \hline
    \end{tabular}
    \vspace*{0.05in}
    \caption{NVIDIA GPU specifications.}
    \vspace*{-0.35in}
    \label{table:gpu_specs}
\end{table}

\parabf{Scalability.}  
Due to various constraints, many organizations operate multiple medium-sized datacenters
instead of one monolithic and giant datacenter.
As training scales out, As training scales out, cross-datacenter training becomes increasingly appealing if it is feasible.
Traditional LLM training, however, involves extensive full-mesh communication operations, 
imposing high bandwidth demands on networking and making cross-datacenter deployment challenging. 
In contrast, the disaggregated RL architecture requires relatively low inter-stage 
communication. Generated samples and related metadata are small in size, 
and although model weights must be transmitted, they only require point-to-point transmission
instead of full-mesh network topology. This is well-suited for 
inter-datacenter dedicated links, making cross-datacenter RL training practically 
feasible rather than merely theoretical. Furthermore, generation instances are entirely 
independent, allowing them to be distributed across multiple datacenters, 
thus fully utilizing the entire resource pool and scaling out.

\subsection{Challenges for Disaggregation}
\label{sec:background:challenges}

\begin{figure}[t!]
    \centering
    \includegraphics[width=0.95\linewidth]{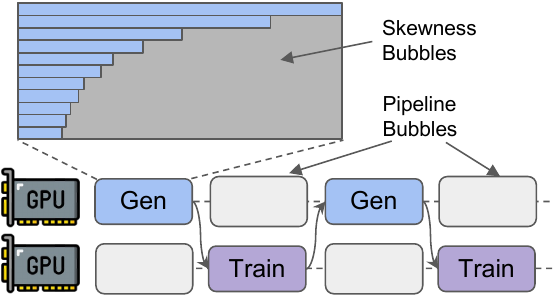}
    \vspace*{-0.1in}
    \caption{Resource waste in disaggregated architecture.}
    \vspace*{-0.2in}
    \label{fig:background:bubbles}
\end{figure}

Although the disaggregated architecture seems natural and promising for the two-stage 
RL workflow, fully unlocking its potential and surpassing the performance of existing 
colocated architectures requires addressing several challenges. 
As shown in Figure~\ref{fig:background:bubbles}, the training timeline under naive disaggregation 
reveals two types of bubbles that lead to GPU under-utilization.

\parabf{Pipeline bubbles.}
This is the primary source of inefficiency in existing disaggregated frameworks. 
The generation stage sends samples to the training 
stage only after all samples have been generated, during which time the 
resources allocated to the training stage remain idle. Similarly, when the 
training stage is active, the generation stage's resources are also left unused,
waiting for the up-to-date model weights. 

Furthermore, traditional LLM training is a relatively static workload, whereas in RL training,
samples are generated online and thus exhibit dynamic behavior.  
As noted in the DeepSeek-R1 technical report~\cite{deepseekr1},
the LLM will spontaneously increase its generation length over time,
enhancing its reasoning ability through self-reflection—an effect referred to 
as \textit{inference-time scaling}~\cite{deepseekr1,openaio1}.
However, the generation and training time respond differently to such workload changes.
As shown on the right side of the Figure~\ref{fig:background:motivation},
with increasing sequence length, generation time grows more significantly than 
that of training. This is primarily due to the enlarged key-value cache size,
which, to avoid out-of-memory (OOM), forces smaller batch sizes during generation
and consequently reduces GPU utilization. To better address pipeline bubbles under 
a disaggregated architecture, it is desirable for the execution time of the two 
stages to be closely matched so that some overlapping techniques like 
mini-batch pipelining~\cite{deepcoder2025} and asynchronous pipelining~\cite{deepcoder2025}
can be applied (\S\ref{sec:tackle_pp_bubbles:overlap}).
However, the differing growth in latency under dynamic workloads leads to stage imbalance, introducing new
bubbles.

\parabf{Skewness bubbles.}
Another type of bubbles originates from the RL workload itself.
In the generation stage, the output length has a skewed distribution~\cite{zhong2024rlhfuse}, 
with only a small subset being much longer than the majority. 
As generation proceeds, only a small set of long-tail samples remain in the system.
This undermines GPU utilization because the decoding phase of generation
requires a large batch size—often in the hundreds—to maintain high throughput
for its memory-bandwidth-bound nature.
Making the problem even worse, under the guidance of inference-time scaling~\cite{deepseekr1,openaio1}, 
the output length continues to grow, making the skewness bubbles an urgent problem 
in real-world deployment.

An engineering workaround~\cite{team2025kimi} is to temporarily store partially generated 
long-tail samples in a replay buffer and generate part of them in each iteration.
However, this changes the original output length distribution and may negatively
impact model quality, introducing a trade-off between accuracy and efficiency.
Another solution, built upon the colocated architecture, is adopted 
by RLHFuse~\cite{zhong2024rlhfuse}. 
It compacts the long-tail samples onto a small subset of resources and uses the 
freed-up machines to pre-execute part of the training stage's work—such as KL divergence 
computation and reward derivation—alongside the generation of long-tail samples, 
effectively filling the skewness bubbles.
However, this approach is not applicable in the disaggregated architecture, 
where the two stages are physically separated.

\section{\sysname Overview}
\label{sec:overview}

\begin{figure}[t!]
    \centering
    \includegraphics[width=0.95\linewidth]{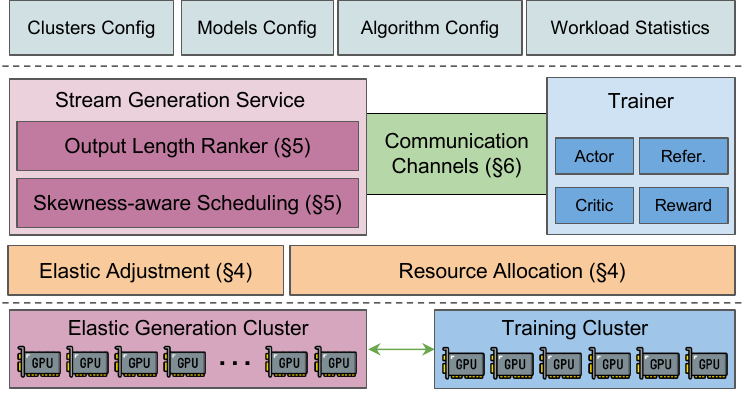}
    \vspace*{-0.1in}
    \caption{\sysname system architecture.}
    \vspace*{-0.2in}
    \label{fig:overview:architecture}
\end{figure}

To this end, we present \sysname, an efficient RL framework
designed with disaggregation from first principle. 
As shown in Figure~\ref{fig:overview:architecture},
\sysname abstracts the generation and training stages
into Stream Generation Service (\SGS) and \Trainer, respectively.
\SGS and \Trainer are deployed on physically separate resources, potentially 
even in different datacenters connected by a point-to-point link.
This architectural design fully unleashes the benefits of disaggregation discussed
 in \S\ref{sec:background:motivation}, enabling (1) \textit{flexible resource allocation},
 (2) \textit{heterogeneous hardware selection}, and (3) \textit{cross-datacenter training}.

We first present a high-level overview of the overall workflow of 
StreamRL.  
Next, we describe in detail our techniques and designs for addressing pipeline  
bubbles (\S\ref{sec:tackle_pp_bubbles}) and skewness bubbles (\S\ref{sec:tackle_longtail}),
as well as the implementation details of the communication between \SGS and \Trainer (\S\ref{sec:implementation}).

\begin{figure*}[t!]
    \centering
    \includegraphics[width=0.95\linewidth]{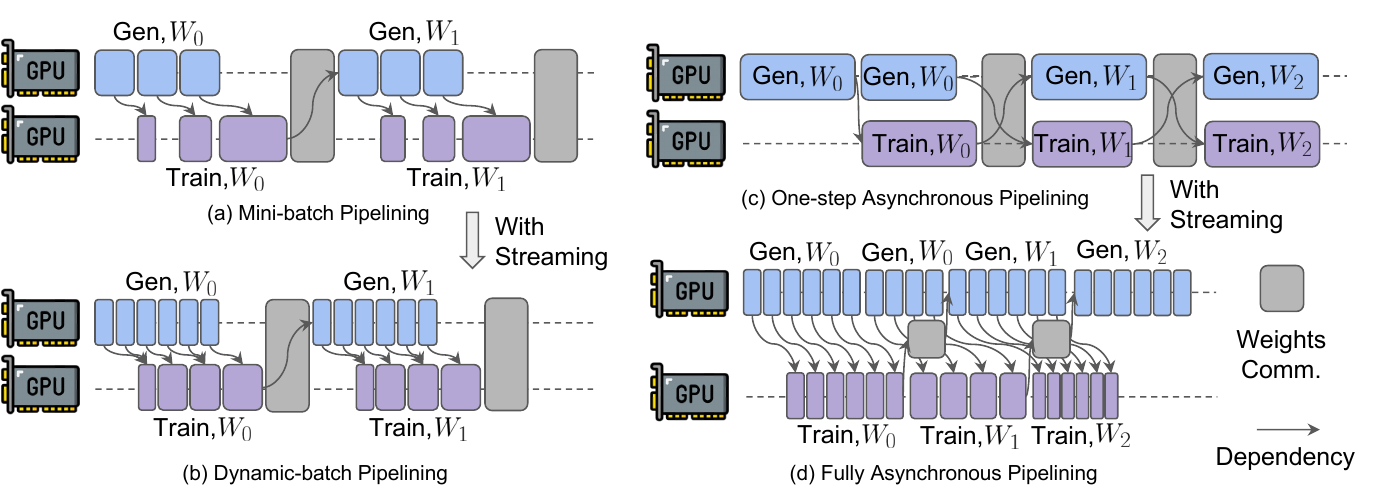}
    \vspace*{-0.1in}
    \caption{How streaming powers existing solutions to better mitigate pipeline bubbles.
    $W_i$ denotes the parameter version.}
    \vspace*{-0.1in}
    \label{fig:background:overlap}
\end{figure*}

\parabf{Workflow.} Given the clusters, models, and algorithm configurations,  
\sysname first determines how to allocate resources between \SGS and \Trainer,
as well as which parallelization strategies to adopt for each.  

During training, \SGS exposes two external
APIs to \Trainer: \verb|update(weights)| and \verb|generate(prompts)|.
\Trainer address pipeline bubbles by adjusting the timing of weights
updates and handling the early streamed-back samples according to the specific RL algorithms.
To address skewness bubbles, \SGS utilizes an output length ranker to identify 
long-tail samples. Based on the predictions, it dispatches prompts to specific
generation instances and decides scheduling order accordingly, 
effectively mitigating the bottleneck caused by long-tail samples.

Although the static configuration ensures the balance between generation and training time 
at the start of training, the dynamic nature of RL workloads requires elastic resource 
adjustment to maintain close execution time between the two stages 
throughout the training process.  
To achieve this, \SGS continuously monitors \Trainer's execution time.
As the workload evolves and sequence length increases, if the generation time 
exceeds the training time by a certain threshold, \SGS automatically scales out 
by increasing its DP size to maintain dynamic balance in execution time.
\section{Tackle Pipeline Bubbles}
\label{sec:tackle_pp_bubbles}

\subsection{Overlapping Design} 
\label{sec:tackle_pp_bubbles:overlap}
To address pipeline bubbles, the key is to ensure that the training stage remains 
active while generation is ongoing. 
Mainstream RL algorithms can generally be categorized into two types: 
synchronous and asynchronous, depending on whether the training samples are generated 
using the latest model weights.  
For each type, we present the current strawman solutions and describe how the 
efficiency of overlapping can be further improved with the support of streaming.

\parabf{Strawman solution 1: Mini-batch pipelining~\cite{deepcoder2025}.}
In synchronous RL, weights update happens after all samples have been processed.  
As shown in Figure~\ref{fig:background:overlap}(a), the samples can be evenly 
divided into several mini-batches analagous to pipeline parallel. 
Once the number of generated samples reaches the size of a mini-batch, they are 
passed to the training stage for processing.  
This approach requires manually setting the mini-batch size: 
if set too large, it reduces the effectiveness of overlapping; 
if too small, it harms training efficiency.  
In practice, the mini-batch size is set empirically to a suitable constant~\cite{deepcoder2025}.
Nevertheless, due to the long-tail effect, the sequence lengths of the later 
mini-batches gradually increase. As a result, the training of the last few 
mini-batches often spill over after generation, creating significant 
pipeline bubbles~\cite{deepcoder2025}. 
Also, due to the imbalance between mini-batches,  
it is hard to set the mini-batch size to avoid idle time in the training stage.

\parabf{Our solution: Dynamic-batch pipelining.} 
We propose replacing the current batched generation 
with \textit{stream generation}, where samples are immediately sent to the training stage 
as soon as completed. This enables sample-level operations such as 
Reference Model inference, KL loss computation, and reward calculation to begin without delay.
As shown in Figure~\ref{fig:background:overlap}(b),  
the training stage can start as soon as it receives enough samples to saturate 
the GPUs, enabling dynamic batching based on the generation speed.  
This eliminates idle time in the training stage except for the first mini-batch
and effectively reduces the bubbles caused by the last few mini-batches.

\parabf{Strawman solution 2: One-step asynchronous pipelining~\cite{deepcoder2025}.}
Essentially, the two stages in synchronous RL still work on the same batch of samples,
so the serialized dependency within each iteration remains, making it impossible 
to achieve perfect overlapping.
Recently, many works~\cite{deepcoder2025, noukhovitch2025asynchronousrlhffasterefficient, wang2025distrlasynchronousdistributedreinforcement, team2025kimi} 
have explored \textit{off-policy asynchronous RL}, where the samples used for training
are not necessarily generated with the up-to-date weights, allowing for some extent of staleness. Existing studies~\cite{noukhovitch2025asynchronousrlhffasterefficient, wang2025distrlasynchronousdistributedreinforcement}
and our experiments~\ref{sec:eval:async} show that one-step asynchronous RL for LLMs does not compromise model performance or convergence.

As shown in Figure~\ref{fig:background:overlap}(c), we can first generate one 
additional batch while the training stage processes samples from the previous 
iteration, thereby shifting the dependency to a cross-iteration manner
and achieving better overlapping.
However, the issue with this batch-level pipelining is that each iteration still 
ends with a global synchronization to transmit the weights, during which 
both stages remain idle. Moreover, due to the dynamic nature of online generation,  
there can be fluctuations in generation and training time across iterations,  
which are difficult to accurately align with resource adjustments, resulting 
in new bubbles.

\parabf{Our solution: Fully asynchronous pipelining.} 
As shown in Figure~\ref{fig:background:overlap}(d),
the above issues can be resolved with streaming.  
First, weight transmission can overlap with the training of the next iteration,  
since samples from the previous iteration have already been streamed and buffered
for training.  
Meanwhile, the generation of the current iteration does not depend on the 
latest weights, and can also proceed in parallel.  
This removes weight transmission completely from the critical path.  
Moreover, even if there are fluctuations in generation and training time across 
iterations, as long as their average speeds are matched and the fluctuation is limited,
no new bubbles will emerge.
Note that here we do not introduce asynchronous samples beyond one step,  
therefore the training semantics remain identical to the naive solution.

\subsection{Stage Balancing}
\label{sec:allocation}

To achieve better overlapping between \SGS and \Trainer,
we need to carefully balance the execution times of the two stages.
As a result, deciding the appropriate parallel strategies and number of GPUs for each stage 
becomes critical for minimizing overall iteration time. 

\begin{algorithm}[t!]
    \caption{Resource Allocation Algorithm.}
    \label{alg:allocation}
    \begin{algorithmic}
    \Require GPU budget, profiler-based estimation model $\mathcal{P}$,
    training workload $\mathcal{W}$.
    \Ensure GPU allocation for \SGS and \Trainer ($x_{\text{opt}}, y_{\text{opt}}$).

    \State \textbf{Single-datacenter: total GPU budget $n$}
    \State $T^* = \infty$
    \For{each configuration $(x, y)$ where $x + y \leq n$}
        \State $T_{\text{gen}} = \mathcal{P}_{\text{gen}}(x, \mathcal{W})$, $T_{\text{train}} = \mathcal{P}_{\text{train}}(y, \mathcal{W})$
        \State $T_{\text{latency}} \gets \max(T_{\text{gen}}, T_{\text{train}})$
        \If{$T_{\text{latency}} < T^*$}
            \State $T^* \gets T_{\text{latency}}, x_{\text{opt}}, y_{\text{opt}} \gets x, y$
        \EndIf
    \EndFor
    \State \Return $x_{\text{opt}}, y_{\text{opt}}$
    \State \textbf{Cross-datacenter: respective GPU budget $m$, $n$}
    \State $T_{\text{gen}} = \mathcal{P}(m, \mathcal{W})$, $T_{\text{train}} = \mathcal{P}(n, \mathcal{W})$
        \If{$T_{\text{gen}} < T_{\text{train}}$}
            \State Find $k$ $\text{s.t.}$ $|\mathcal{P}_{\text{gen}}(k, \mathcal{W}) - T_{\text{train}}|$ achieves minimum.
            \State \Return $k, n$
        \Else
            \State Find $k$ $\text{s.t.}$ $|\mathcal{P}_{\text{train}}(k, \mathcal{W}) - T_{\text{gen}}|$ achieves minimum.
            \State \Return $m, k$
        \EndIf
    \end{algorithmic}
\end{algorithm}

\parabf{Parallel configuration.}
Before deciding how many resources to allocate to each stage, we first address a subproblem: 
determining the optimal execution time of either \SGS or \Trainer under a given workload 
and GPU budget. This essentially reduces to optimizing the parallel strategy, 
which is a well-studied problem for both LLM training~\cite{alpa,zhang2024disttrain} and generation~\cite{zhong2024distserve}. 
For \Trainer, we adopt a profiler-based approach inspired by prior work on automated 
parallelism~\cite{alpa, wang2019supporting, zhang2024disttrain, miao2022galvatron}.
Due to the determinism of DNN execution time~\cite{gujarati2020serving}, we can accurately 
model training time under a fixed GPU budget with minimal profiling. 
For \SGS, generation time depends on the scheduling strategy during 
inference~\cite{li2023alpaserve}. Fortunately, under our skewness-aware 
scheduling (\S\ref{sec:scheduling:longtail}), generation time is also deterministic 
for a given workload, allowing us to model it similarly. 
Note that we assume access to the RL workloads. In practice, this can be obtained 
from samples generated by recent training iterations or bootstrapped from samples generated 
by the LLM prior to training.

\parabf{Resource allocation.}
Building on the above strategies, we can determine the resource allocation 
for each stage. \sysname supports two deployment solutions in production RL training.
The first is single-datacenter deployment, where \SGS and \Trainer are colocated within the
same datacenter equipped with homogeneous hardware resources.
This is the standard setup in prior LLM training systems~\cite{shoeybi2020megatronlm,jiang2024megascale}.
Also, \sysname supports cross-datacenter deployment, leveraging the decoupled nature of \SGS and \Trainer 
in RL workflows and placing \SGS and \Trainer in separate datacenters with heterogeneous hardware (e.g., H20 vs. H800).
We present the resource allocation algorithms under two different deployments as shown in Algorithm~\ref{alg:allocation}.

\parait{\underline{Single-datacenter.}}
We define the number of GPUs allocated to \SGS and \Trainer as $x$ and $y$, respectively.
The resource constraint is $x + y \leq n$, where $n$ denotes the total GPU budget.
To determine the optimal allocation strategy, we enumerates
the allocation configurations. For each case, we can get the generation and 
training time respectively using the aforementioned profiler-based modeling.
Then we use the larger of the two as the estimated latency
and selects the optimal $(x, y)$ pair that minimizes overall iteration time.

\parait{\underline{Cross-datacenter.}}
For cross-datacenter deployment, let $m$ and $n$ denote the available GPUs 
in the datacenters for \SGS and \Trainer deployment, respectively.
The resource constraints are $x \leq m$ and $y \leq n$, making $x$ and $y$ independent variables.
A naive choice is $(x, y) = (m, n)$, which fully utilizes all the resources in both datacenters.
However, since iteration time is determined by the slower stage of \SGS and \Trainer, 
such full allocation may lead to resource waste.
To address this, we identify the faster stage under full allocation and gradually reduce 
its GPU usage until both stages achieve similar execution times.
This strategy eliminates unnecessary GPU usage, allowing surplus resources to be 
reallocated to other jobs within the respective datacenter.

\parabf{Dynamic adjustment.}
The above techniques only ensure that the two stages are balanced at the beginning of training.
As observed in the DeepSeek-R1 technical report~\cite{deepseekr1}, the generation length of LLMs
increases progressively during RL training, leading to evolving compute and memory demands for 
both \SGS and \Trainer stages. Unfortunately, the two stages exhibit different sensitivities to 
workload changes in terms of latency.
To address this, we propose a dynamic adjustment mechanism that monitors the execution time
gap between generation and training, denoted as $\delta$.
As shown in Figure~\ref{fig:background:motivation}, generation time increases faster than training,
which means $\delta$ will gradually increase as training proceeds.

Ideally, when $\delta$ exceeds a certain threshold, we can rerun the resource allocation 
algorithm to rebalance the stages. However, in practice, all the GPUs in
Trainer are tightly coupled through communication group due to the 3D parallelism.
Changing the parallel strategy or reallocating resources for \Trainer requires restarting 
the entire training runtime, which incurs significant overhead. 
In contrast, generation instances in \SGS are naturally decoupled.
Therefore, \sysname estimates the reduction in generation time, $\delta'$, achievable by 
adding one data parallel (DP) unit to \SGS. $\delta'$ is calculated using the aforementioned 
profiler and the current RL workload. When $\delta \geq \delta'$, adjustment is triggered by 
adding one more DP unit to \SGS. This reallocation does not interrupt training, 
and the overhead is limited to initializing the added DP unit, which is 
negligible relative to the overall RL training time.

\section{Tackle Skewness Bubbles}
\label{sec:tackle_longtail}

In this section, we introduce the techniques used by \SGS to tackle skewness
bubbles. At a high-level, \SGS minimizes the generation time under given resources.
which serves as a function used by the resource allocation section (\S\ref{sec:allocation}).

\begin{figure}[t!]
    \centering
    \includegraphics[width=1.0\linewidth]{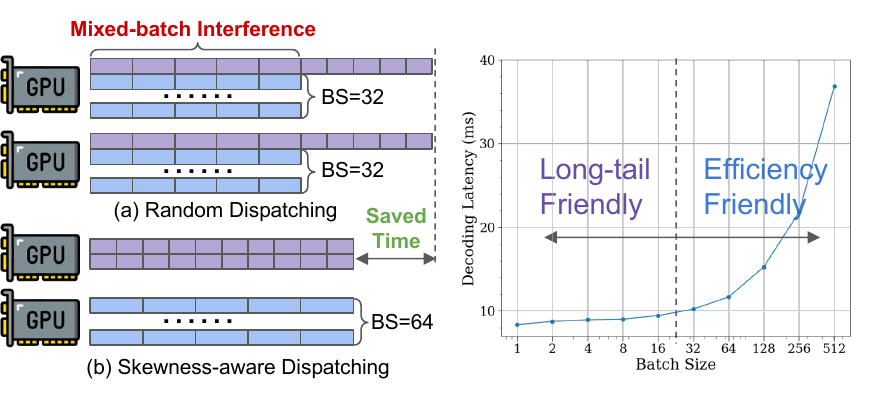}
    \vspace*{-0.3in}
    \caption{Left: The advantage of skewness-aware dispatching over random dispatching.
    Right: The trend of per-token decoding latency for a 7B LLM profiled on NVIDIA H800 
    with vLLM~\cite{vllm} as the batch size increases.}
    \vspace*{-0.2in}
    \label{fig:design:scheduling}
\end{figure}

\subsection{Problems and Opportunities}
\parabf{Problem 1.} Existing systems do not differentiate between long-tail samples and regular samples.
To achieve workload balance, prompts are typically assigned randomly across generation 
instances. Figure~\ref{fig:design:scheduling}(a) shows
an simple example where the generation DP size is 2, with 2 long-tail samples whose output length
are two times that of the remaining 64 samples. Under the random dispatching strategy, 
each generation instance receives one long-tail sample along with half of the regular 
samples. This results in significant interference for the long-tail samples during the 
first half of generation due to batched inference.

The right side of Figure~\ref{fig:design:scheduling} shows the trend of per-token 
decoding latency for a 13B model on an NVIDIA A100 GPU as the batch size increases. 
It can be observed that latency grows slowly before reaching compute-bound, 
and then increases almost linearly after that. To improve throughput, 
existing systems usually accumulate a sufficiently large batch size for each 
instance through random dispatching. However, in the presence 
of long-tail samples, this approach not only slows down their decoding in the early 
stages but also leads to extremely low utilization in the later stage,
as only a few long-tail samples remain in the system.

\parabf{Opportunity 1.} 
With an intuitive understanding of the problem, we now proceed to the solution.
The generation latency of a sample can be modeled as:
\begin{equation}
    Sample \ Latency = PTL(BS) \times L
    \label{eq:sample_latency}
\end{equation}
where per-token latency ($PTL$) is a function of batch size ($BS$) which can be profiled
in advance and $L$ is the output length.
It can be observed that the random dispatching strategy
balances load merely based on $L$, without considering $PTL$.  
For samples with longer $L$, we actually prefer to reduce their $PTL$—that is, 
their $BS$—since $PTL$ is a monotonically increasing function of $BS$.

The heuristics for the solution naturally emerge: we can extract the long-tail samples 
and assign them to a few dedicated instances with smaller batch sizes, allowing them to 
decode at the fastest possible speed and eliminating interference caused by batching.  
Meanwhile, regular samples can be grouped into large batches to fully utilize GPU resources.  
By extending the original one-dimensional load balancing into a two-dimensional scheme, 
generation latency can be effectively reduced, as illustrated in Figure~\ref{fig:design:scheduling}(b).

\parabf{Problem 2.} The above approach relies on a key assumption: that long-tail samples 
can be identified before generation begins. However, the output length of LLM generation 
is typically regarded as \textit{not known a priori}.

\parabf{Opportunity 2.}
Fortunately, while it is difficult to predict the exact generation length of each sample,
it is possible to estimate the \textit{relative ranks} of output lengths~\cite{fu2024efficientllmschedulinglearning}
with another model.  
Intuitively, the ranking problem is essentially a classification problem, as more difficult
prompts typically require more reasoning—i.e., longer output lengths. 
Therefore, the ranking model essentially classifies prompts based on their difficulty.  
Difficulty is a property inherent to the prompts themselves, which can be generalized 
across different LLMs, leading to relatively high prediction accuracy.  
In contrast, the exact output length is a characteristic of the LLM itself and is 
significantly more challenging to predict.  
Our experiments (\S\ref{sec:eval:ablation}) show that the top 20\% long-tail samples
can be recalled with nearly 90\% accuracy, which aligns well with our hypothesis.  
Since the generation time is predominantly bottlenecked by the long-tail samples,
accurately identifying them is sufficient to achieve the majority of the performance gains
compared to the upper bound speedup where the oracle is available (\S\ref{sec:eval:ablation}).

Next, we detail how the above observations are incorporated into the design of 
the output length ranker (\S\ref{sec:scheduling:predictor}) and the skewness-aware 
scheduling (\S\ref{sec:scheduling:longtail}).

\subsection{Output Length Ranker}
\label{sec:scheduling:predictor}

\parabf{Method.} 
To train the ranker model, we collect a set of input prompts along with their 
corresponding output lengths from the target LLM. These (prompt, length) pairs can be 
sourced from our online inference serving service or generated offline before training.
We then concatenate the prompts with their corresponding output lengths to form a training
dataset. Using this dataset, we directly perform supervised fine-tuning (SFT) on a 
small LLM as the ranker model.

After fine-tuning, the ranker model can take a batch of prompts as input and estimate
their absolute output lengths. These estimated lengths are then used to sort the prompts,
producing the final ranking result.
Note that the SFT process involves predicting absolute lengths. As RL training progresses 
and the parameters of the target LLM evolve, even the same prompt may yield different 
output lengths. Therefore, after a period of training, we perform online fine-tuning 
of the ranker model using recent generation results, following the same methodology.

Fortunately, the difficulty of a prompt is an inherent property and remains stable. 
As a result, even if the absolute predictions drift, the relative ranking produced 
by the ranker remains reasonably accurate, reducing the need for frequent online fine-tuning.

\parabf{Overhead.}
One concern is the overhead introduced by the ranker model.  
First, the ranker model trains very quickly, requiring only a few minutes to converge.  
After training, we perform a one-time offline preprocessing step on the dataset used for RL,
estimating the actual output length for each prompt. These estimates serve as the basis for 
the subsequent skewness-aware scheduling. Note that this preprocessing is conducted entirely 
offline, meaning the ranker model imposes no online overhead on the RL training process 
and does not affect the original training efficiency.

\subsection{Skewness-aware Scheduling}
\label{sec:scheduling:longtail}

\begin{algorithm}[t!]
    \caption{Skewness-aware Dispatching Algorithm.}
    \label{alg:dispatch}
    \begin{algorithmic}
    \Require Batch of prompts $\mathcal{P}$, estimated lengths $\mathcal{L}$, longtail threshold $\alpha$, output length distribution $\mathcal{D}$, and $N$ generation instances
    \Ensure Number of DP instances for long-tail and regular samples respectively.
    \State $\mathcal{P} \gets \text{Sort}(\mathcal{P}, \mathcal{L}, \text{descending})$
    \State $\mathcal{P}_\alpha \gets \mathcal{P}[:\alpha \times |\mathcal{P}|]$ \Comment{Long-tail samples}
    \State $\mathcal{P}_r \gets \mathcal{P}[\alpha \times |\mathcal{P}|:]$ \Comment{Regular samples}
    \State $L_{\alpha} \gets \text{P90}(\mathcal{D})$, $L_r \gets \text{P50}(\mathcal{D})$, $L^* \gets \infty$
    \For{$N_l, N_r$ such that $N_l + N_r = N$}
        \State $L_{\text{total\_latency}} \gets L_{\text{latency}}(\mathcal{P}_\alpha, L_{\alpha}, N_l) + L_{\text{latency}}(\mathcal{P}_r, L_r, N_r)$
        \If{$L_{\text{total\_latency}} < L^*$}
            \State $L^* \gets L_{\text{total\_latency}}$, $N_l^* \gets N_l$, $N_r^* \gets N_r$
        \EndIf
    \EndFor
    \State \Return ($N_l^*$, $N_r^*$)
    \end{algorithmic}
\end{algorithm}

With the help from output length ranker, \SGS will receive a batch 
of prompts along with their estimated output lengths. 
It then needs to make two decisions: 
determine how to dispatch the prompts to different generation instances,
and decide the scheduling order within each instance after dispatching.

\parabf{Dispatching.}
\label{sec:scheduling:distributor}
Given a batch of prompts, we first sort them by their estimated output lengths 
from longest to shortest. With the relative order established, we mark the 
longest $\alpha\%$ of them as long-tail samples, where $\alpha$ is a 
hyperparameter; in practice, setting $\alpha$ to 20 yields good results (\S\ref{sec:eval:async}). 
Next, we need to select $N_l$ out of $N$ generation instances to handle the 
long-tail samples, while the remaining $N_r$ instances are used for regular samples.

To ensure workload balance, we need an estimate of the actual workload for 
the regular and long-tail samples.
Here, we assume access to the output length distribution $\mathcal{D}$ of the LLM to 
be trained. This workload characteristic can be derived from recently generated samples 
during training or by having the LLM generate a set of samples beforehand for 
bootstrapping. We use the P50 and P90 of $\mathcal{D}$ to estimate the average 
output lengths for regular and long-tail samples, respectively.
Based on this, we extend the sample latency (\ref{eq:sample_latency}) to estimate 
the single instance generation latency:
\begin{equation}
    Latency = PTL(BS) \times L_{avg} \times \lceil \frac{M}{BS} \rceil
    \label{eq:generation_latency}
\end{equation}
where $L_{avg}$ is the estimated average output length of the samples assigned to the 
instance, depending on whether the instance processes regular or long-tail samples, 
and $M$ is the number of prompts assigned to the instance. Ideally, we would 
like $BS=M$, so that all samples can be processed in a single round. However, 
with longer output lengths, a larger $BS$ will result in higher key-value cache 
memory usage, so $BS$ is constrained by the GPU memory capacity.

With Equation~\ref{eq:generation_latency}, we can iterate 
through all $(N_l, N_r)$ configurations and find the one that minimizes the generation time.
Algorithm~\ref{alg:dispatch} shows the pseudocode for the skewness-aware dispatching algorithm.
After that, the long-tail and regular samples 
can be evenly distributed within their respective instances.

\parabf{Scheduling Order.}
\label{sec:scheduling:order}
As mentioned earlier, $BS$ is limited by key-value cache memory usage, so the number of 
samples $M$ assigned to each instance may exceed the $BS$ used during generation. 
In this case, multiple rounds of generation are required, which introduces the need for 
deciding the scheduling order of samples. 
This problem is a variant of the $P||C_{\max}$ (makespan minimization) problem, where $P$ denotes 
parallel processing units and $C_{\max}$ is the maximum completion time. In our case, each generation 
iteration of batch size $BS$ is viewed as $BS$ parallel processing units. 
We employ a well-known greedy algorithm, longest-processing-time-first (LPT) 
scheduling~\cite{graham1969bounds}, to address this problem. Specifically, samples are assigned to 
the batch in descending order of their estimated output lengths. Once a sample is completed, 
the sample with the longest remaining output length is added to the batch. 
This process continues until all samples are processed. Prior work~\cite{graham1969bounds} has proved 
that LPT scheduling is $4/3$-approximation, i.e., its completion time is at most $4/3 \times$ that of the 
optimal scheduling.

\section{Implementation}
\label{sec:implementation}
\parabf{RL Training Framework.} \SGS employs an in-house inference engine implemented in C++
with optimized CUDA kernels, supporting continuous batching~\cite{yu2022orca} to release shorter samples 
early and prefix sharing~\cite{zheng2024sglangefficientexecutionstructured} to save key-value cache usage.
\Trainer implements 3D parallelism similar to prior work~\cite{shoeybi2020megatronlm,jiang2024megascale,mei2024realhf}.
To address GPU memory constraints, we develop dynamic CPU offloading that interleaves 
the execution of different models through memory swapping.

\parabf{Tensor-native RPC Library.}
Conventional distributed computing frameworks~\cite{ray, torchrpc} typically incur significant serialization 
and deserialization costs for tensor data transfers. 
We developed RL-RPC, a communication framework optimized for efficient data transfer between 
\SGS and \Trainer. The system employs GPU-Direct RDMA for zero-copy tensor transfers, 
bypassing CPU involvement and eliminating serialization overhead to minimize communication costs.
By fully leveraging RDMA bandwidth without consuming GPU SM resources, RL-RPC prevents performance 
degradation through overlapping of communication and computation. A TCP fallback mechanism ensures compatibility
across different network environments, including non-RDMA cross-datacenter connections.

\parabf{Weights transmission.}
After trainer-side weights sharding, \sysname employs a network-aware transmission engine 
to efficiently broadcast weights from \Trainer to \SGS. The engine dynamically builds 
broadcast trees optimized for network topology. In the single-datacenter setting,
where both reside on the same RDMA network, it creates multiple trees rooted at different 
DP ranks, load-balancing across trees to keep all DPs bandwidth-saturated. 
For cross-datacenter deployment with limited connection bandwidth,
only the root (DP rank 0) sends weights to a desinated \SGS DP instance in the remote datacenter, 
followed by a local broadcast to minimizes cross-datacenter traffic.

\begin{table}[t!]
    \centering
    \resizebox{0.8\linewidth}{!} {
    \begin{tabular}{lccccc}
        \toprule
        \multirow{2}{*}{\textbf{Models}} & \textbf{\# of} & \textbf{\# of} & \textbf{\# of} & \textbf{Hidden} \\
                                         & \textbf{Layers} & \textbf{Q Heads} & \textbf{K/V Heads} & \textbf{Size} \\
        \midrule
        Qwen2.5-7B  & 28 & 28 & 8 & 3584 \\
        Qwen2.5-32B & 64 & 40 & 8 & 5120 \\
        Qwen2.5-72B & 80 & 64 & 8 & 8192 \\
        \bottomrule
    \end{tabular}
    }
    \caption{LLM specifications.}
    \vspace{-0.3in}
    \label{eval:tab:configs}
\end{table}

\begin{figure}[t!]
    \centering
    \includegraphics[width=0.95\linewidth]{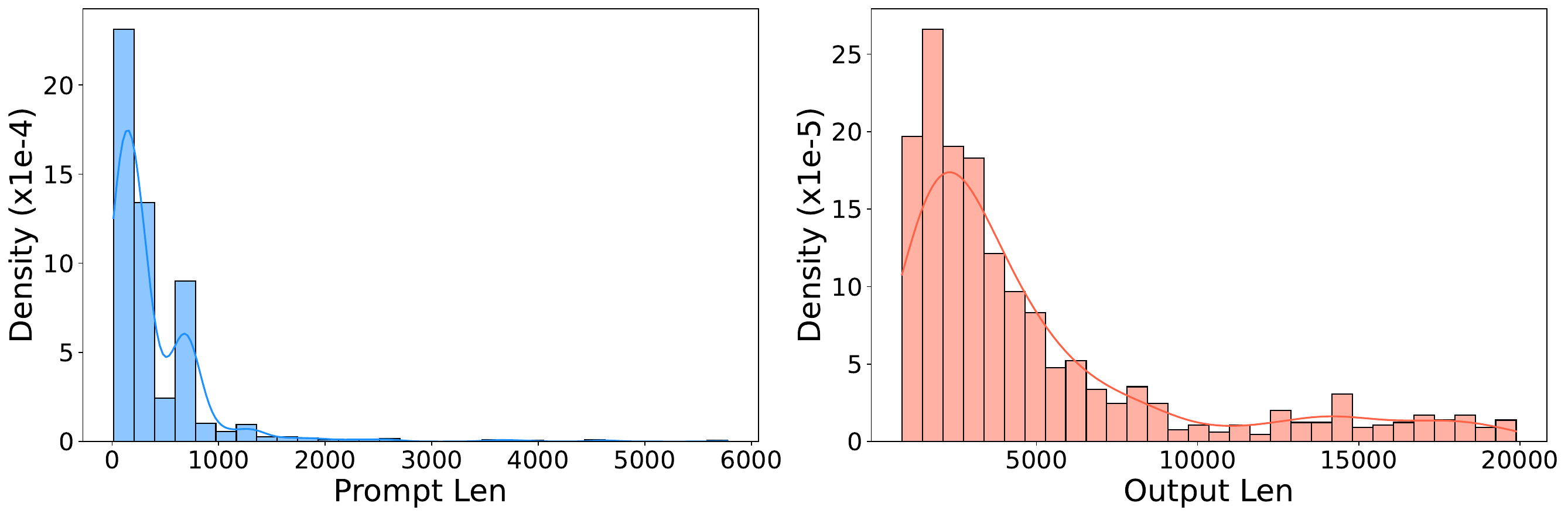}
    \vspace*{-0.2in}
    \caption{The prompt and output length distribution of the evaluation dataset.}
    \vspace*{-0.2in}
    \label{fig:eval:distribution}
\end{figure}

\begin{figure*}[t!]
    \centering
    \includegraphics[width=0.95\linewidth]{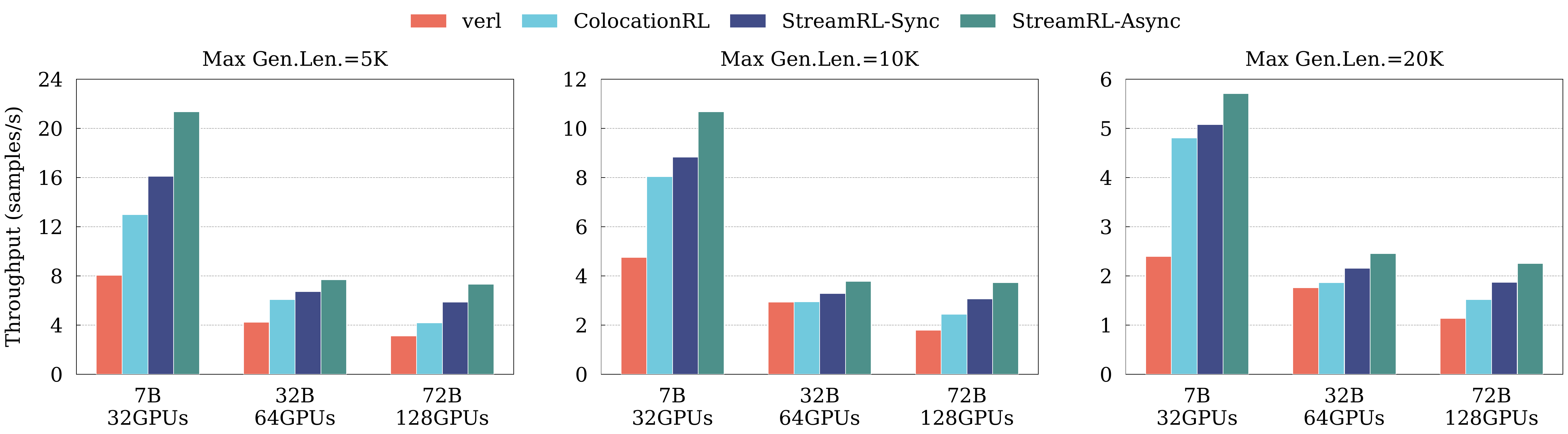}
    \vspace*{-0.1in}
    \caption{End-to-end throughput of RL training systems under different
    sequence length and model size settings.}
    \vspace*{-0.1in}
    \label{fig:eval:end2end}
\end{figure*}

\section{Evaluation}
\label{sec:evaluation}

In this section, we evaluate \sysname with LLMs of different sizes ranging from
7B to 72B on real-world dataset. First, we compare the end-to-end performance
of \sysname to other RL training frameworks under single-datacenter setting (\S\ref{sec:eval:end2end}),
and conduct ablation studies to show the effectiveness of our proposed techniques (\S\ref{sec:eval:ablation}).
Next, we show the performance of \sysname under heterogeneous, cross-datacenter setting to
demonstrate the flexibility and scalability of disaggregated architecture (\S\ref{sec:eval:heterogeneity}).
Finally, we provide training curves to demonstrate that asynchronous RL achieves comparable performance and 
convergence to synchronous RL (\S\ref{sec:eval:async}).

\parabf{Testbed.} We deploy \sysname on a H800 cluster with 16 nodes and 128 GPUs.
Each node has 8 NVIDIA H800-80GB GPUs. Nodes are
connected by 8 * 200 Gbps RDMA network based on RoCEv2 with rail-optimized topology.
For the heterogeneous and cross-datacenter experiments (\S\ref{sec:eval:heterogeneity}),
we also utilize a cloud-based H20 cluster with 4 nodes and 32 GPUs. Each node
has 8 NVIDIA H20-96GB GPUs. Nodes are connect by 100Gbps TCP network.
The H800 and H20 clusters are connected by a 80Gbps dedicated link.
Other specifications between H800 and H20 are listed in Table~\ref{table:gpu_specs}.

\parabf{Models.} We choose Qwen2.5 models~\cite{qwen} ranging from 7B to 72B,
which is a popular base model family used for RL post-training both in academia and industry.
The detailed model architectures are listed in Table~\ref{eval:tab:configs}. 

\parabf{Dataset.} We use an internal CodeMath prompts dataset and collect responses from DeepSeek-R1~\cite{deepseekr1},
an open-source, advanced reasoning model, as ground truth.
The distributions of prompt length and output length in the dataset are shown in Figure~\ref{fig:eval:distribution}.
The maximum output length is 20K and the distribution is very long-tail.
To train \sysname's output length ranker model, we split the dataset into training, 
validation, and test sets with a ratio of 7:2:1. All performance evaluations are 
conducted on samples from the test set.
To avoid discrepancies in generation length caused by numerical differences in 
underlying runtime across different RL frameworks,
we modify the inference code in all RL frameworks to generate outputs with the 
same length following the ground truth of each prompt. This not only help simulating 
the long-tail effects observed in real-world RL training while ensuring
a fair comparison across different RL frameworks.

\parabf{Settings.} We use the PPO algorithm~\cite{schulman2017proximal}
which is widely used in RL post-training. But note that the effectiveness of \sysname does not rely on 
any specific RL algorithm and can generalize to others such as GRPO~\cite{shao2024deepseekmath}. 
We set the Actor, Critic, and Reference Model to the same size.
We do not use an explicit Reward Model but a rule-based verifier to provide rewards following~\cite{deepseekr1}.
We also proportionally scale down the output lengths by two and four,
to simulate the early and middle stages of RL training, when the output lengths are not yet particularly long. 
This leads to three datasets, which we denote as 5K, 10K, and 20K for clarity.
In each iteration, we use a global batch size of 1024 following~\cite{sheng2024hybridflow}.

\parabf{Metrics.} For the end-to-end experiment, we measure the sample throughput
following~\cite{zhong2024rlhfuse}, which is defined as 
the average number of samples processed per second. Under each setting, we 
record the sample throughput over 20 consecutive training iterations after warm-up.

\subsection{End-to-end Experiments}
\label{sec:eval:end2end}

We compare the end-to-end performance of \sysname against the following
baseline frameworks.
\begin{itemize}[leftmargin=*]
    \item verl~\cite{sheng2024hybridflow} is the state-of-the-art open-source RL training 
    framework and a representative of the colocated architecture.
    It proposes a hierarchical hybrid programming model for the RL dataflow 
    and optimizes the parallel strategies of each model. 
    We choose vLLM~\cite{kwon2023efficient} as its inference engine and 
    Megatron-LM~\cite{shoeybi2020megatronlm} as its training backend.
    \item ColocationRL is our in-house RL training framework based on 
    a colocated architecture. It shares the same inference and training backend implementations 
    as \sysname. 
    We include this baseline to demonstrate the performance improvements 
    brought by disaggregation and our techniques in \S\ref{sec:tackle_pp_bubbles} 
    and \S\ref{sec:tackle_longtail}, eliminating any unfair comparisons caused by differences 
    in underlying implementations and other optimization techniques in LLM generation and training 
    that are orthogonal to our core contributions.
\end{itemize}
We do not compare against other open-source frameworks like OpenRLHF~\cite{hu2024openrlhf} 
and NeMo~\cite{Harper_NeMo_a_toolkit} which are based on disaggregated architectures, 
as they have lower throughput than verl due to resource idleness (\S\ref{sec:background:problems}) 
as reported in~\cite{sheng2024hybridflow}.
For \sysname, we show two variants. \sysname-Sync implements the synchronous version of PPO,
which is the same as the baselines. \sysname-Async implements the one-step asynchronous version
to maximize throughput.

Figure~\ref{fig:eval:end2end} presents the end-to-end throughput of various RL frameworks under 
different maximum sequence lengths and model sizes. Compared to verl, \sysname-Sync achieves 
a $1.12\times$--$2.12\times$ speedup, partially attributed to optimizations in the underlying 
inference and training framework. Compared to ColocationRL, \sysname-Sync 
achieves a $1.06\times$--$1.41\times$ speedup by leveraging disaggregated stream generation and
skewness-aware scheduling.
Under the Colocation setup, generation is highly memory-bandwidth-bound, leading to low GPU utilization.
In contrast, disaggregation enables \sysname-Sync to flexibly and judiciously allocate resources
for generation and effectively overlaps the two stages via streaming, thereby improving GPU utilization.
However, even with streaming, the performance gains from disaggregation are partially offset due to 
the long-tail distribution of data and stage dependencies. \sysname-Async further addresses these 
limitations by employing one-step asynchronous training to fully overlap pipeline bubbles, 
achieving $1.30\times$--$2.66\times$ throughput improvement.

\subsection{Ablation Studies}
\label{sec:eval:ablation}

\begin{table}[t]
    \centering
    \begin{tabular}{|c|c|c|}
        \hline
        \multirow{2}{*}{\centering Idx} & \multirow{2}{*}{\centering Method} & Normalized \\
            &        & Throughput \\
        \hline
        1 &  Colocation Baseline & 1.00 \\
        2 &  (1) with skewness-aware scheduling & 1.08 (+8\%)  \\
        3 &  (2) with disaggregation + streaming & 1.23 (+15\%) \\
        4 &  (3) with asynchronous & 1.48 (+25\%) \\
        \hline
    \end{tabular}
    \vspace*{0.05in}
    \caption{Throughput improvement breakdown when training 72B model on the 20K dataset.}
    \vspace*{-0.2in}
    \label{tab:exp-ablation}
\end{table}

\parabf{Improvement breakdown.} Table~\ref{tab:exp-ablation} shows
the detailed improvement breakdown of our proposed techniques 
on the 72B model under the dataset with 20K maximum length. 

We observe that (2) skewness-aware scheduling improves throughput by 8\% over 
ColocationRL, primarily by optimizing generation time. Using the output length 
ranker model to identify long-tail samples, we accelerate their generation by 
assigning them dedicated compute resources and smaller batch sizes. The effectiveness of 
this skewness-aware scheduling depends on the prediction accuracy of the ranker model 
for long-tail samples. Table~\ref{tab:prediction-accuracy} shows the recall rates for 
different proportions of long-tail samples using models trained with base models of varying sizes.
For the longest 20\% of samples, we can achieve the recall rate up to 87\%. 

\begin{table}[t]
    \centering
    \begin{tabular}{|c|c|c|c|}
        \hline
        Base Model & Tail 20\% & Tail 10\% & Tail 5\% \\
        \hline
        Qwen2.5-7B & 0.87 & 0.82 & 0.76 \\
        \hline
        Qwen2.5-3B & 0.85 & 0.79 & 0.72 \\
        \hline
        Qwen2.5-1.5B & 0.81 & 0.75 & 0.68 \\
        \hline
    \end{tabular}
    \vspace*{0.05in}
    \caption{Recall rate under different tail rates and base models.}
    \vspace*{-0.2in}
    \label{tab:prediction-accuracy}
\end{table}

\begin{figure}[t!]
    \centering
    \includegraphics[width=0.95\linewidth]{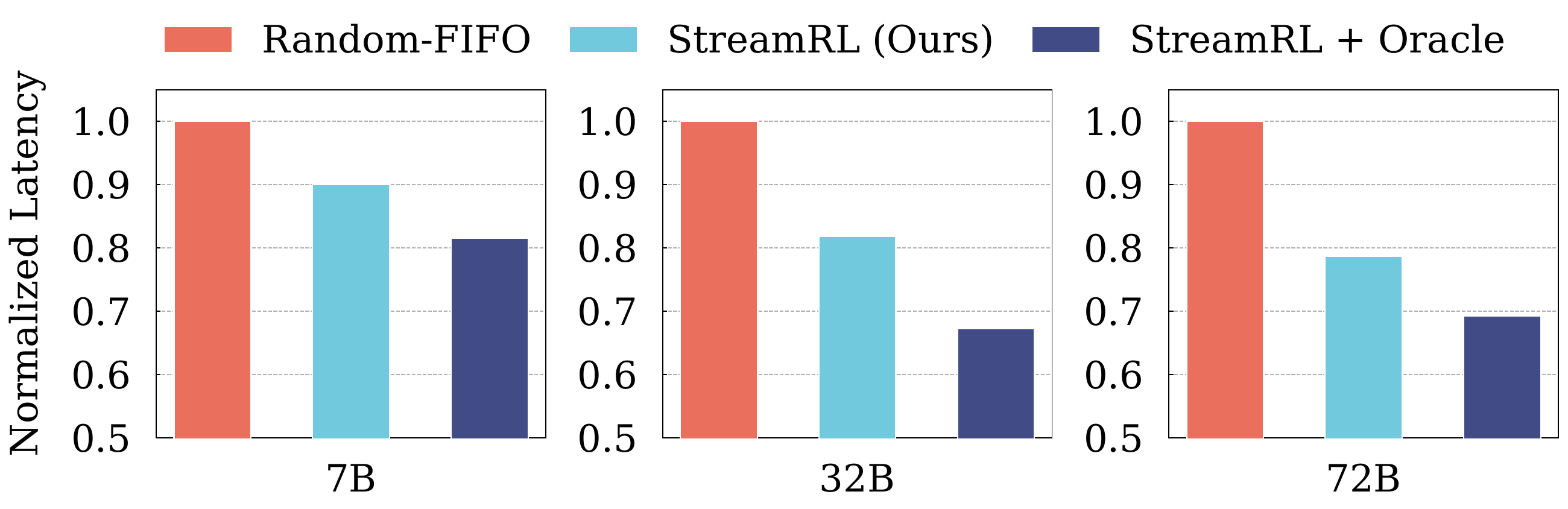}
    \vspace*{-0.1in}
    \caption{Generation time with different scheduling algorithms on various models 
    under the 20K dataset.}
    \vspace*{-0.1in}
    \label{fig:eval:scheduling}
\end{figure}

How much performance is impacted by the remaining unpredicted long-tail samples?
We compare generation time under random dispatching and skewness-aware scheduling, 
and also evaluate an oracle setting where output lengths are known in advance as the speedup upper bound.
As shown in Figure~\ref{fig:eval:scheduling}, we achieve most of the potential gains with our ranker model.
As RL training progresses, the output length distribution of the LLM evolves.
To maintain prediction accuracy, we periodically perform online finetuning of the ranker model using 
recently generated samples to adapt to the distribution shift. The convergence time for this 
training is just a couple of minutes, which is negligible compared to the overall RL training time.

Building on skewness-aware scheduling, (3) disaggregated streaming further improves throughput by 15\%,
and (4) asynchronous training yields an additional 25\% gain. These improvements fundamentally stem 
from the ability of disaggregation to allocate appropriate resources to the generation stage, 
thereby increasing its GPU utilization. To convert this reclaimed utilization into end-to-end speedup,
it is necessary to address pipeline bubbles caused by stage dependencies. Streaming overlaps part of 
the bubbles, while asynchronous training achieves nearly the full overlapping.

\begin{figure}[t!]
    \centering
    \includegraphics[width=0.95\linewidth]{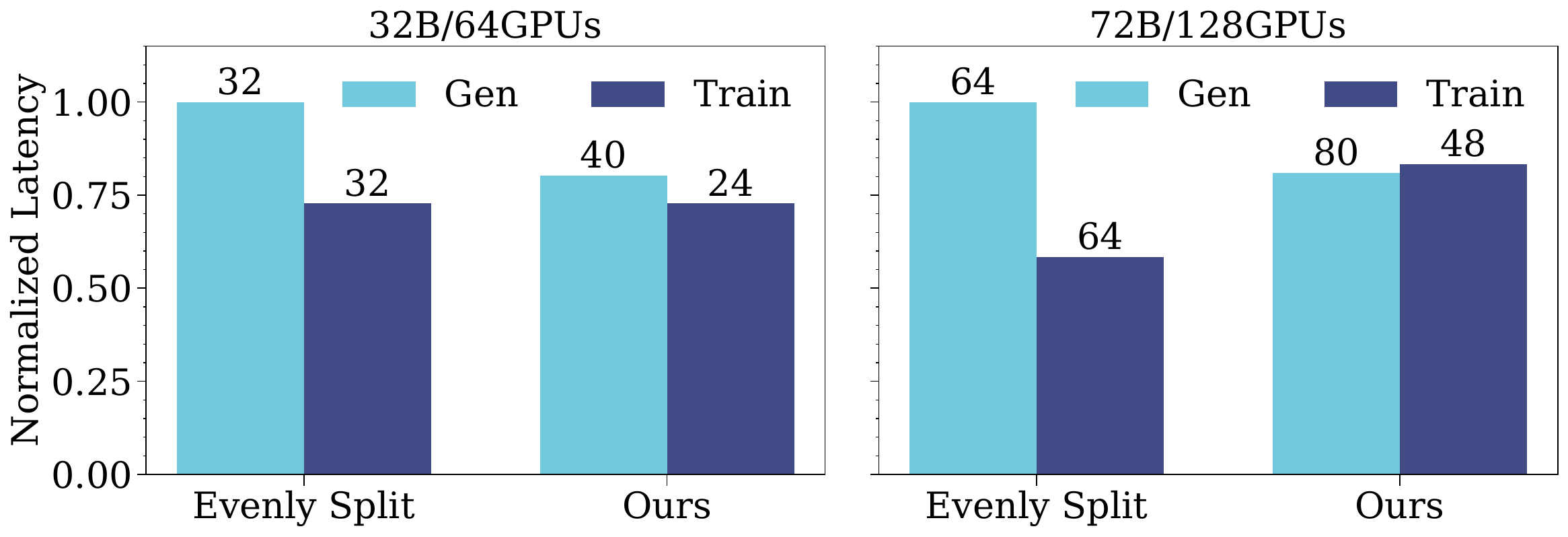}
    \vspace*{-0.1in}
    \caption{The iteration time breakdown compared between even resource split and
    our resource allocation algorithm when training 32B and 72B model on the 20K dataset.}
    \vspace*{-0.1in}
    \label{fig:eval:static}
\end{figure}

\parabf{Resource allocation.} To achieve better overlapping, balancing the latency of the 
two stages is also critical. Figure~\ref{fig:eval:static} compares the iteration time breakdown
under a naive evenly-split scheme and the ones selected by \sysname's 
resource allocation algorithm. The number of GPUs used for each stage is annotated 
at the top of each bar. As shown, by adjusting the parallel strategies and 
resource allocation, we achieve well-balanced stage latencies. In asynchronous training, 
the iteration time is determined by the slower of the two stages, so balanced stage latencies 
directly translate into the speedup of $1.25\times$.

\begin{figure}[t!]
    \centering
    \includegraphics[width=0.95\linewidth]{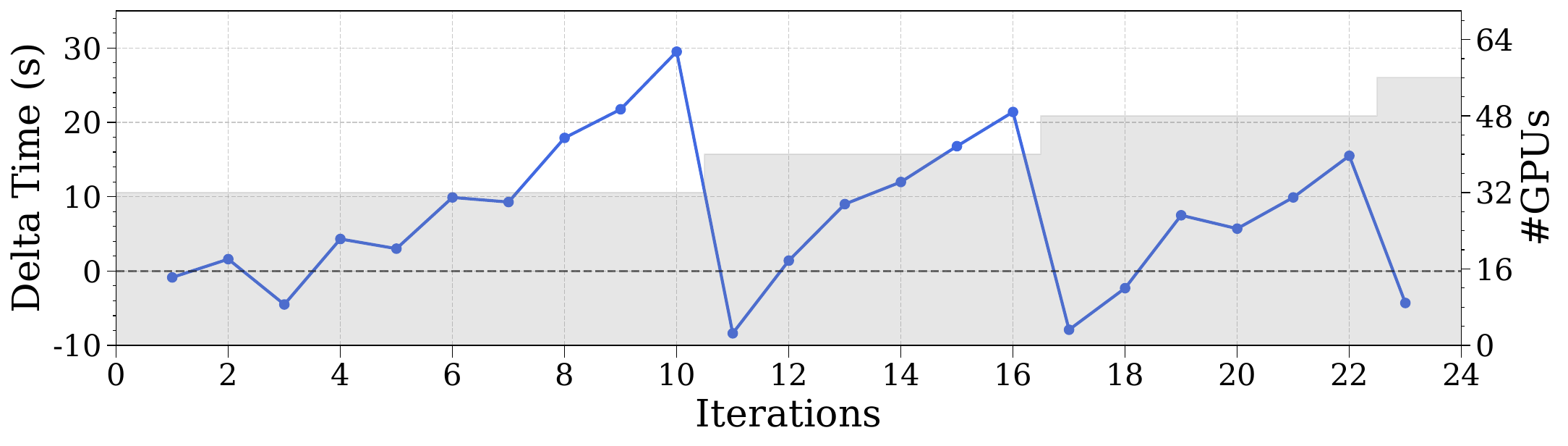}
    \vspace*{-0.15in}
    \caption{The delta time between the two stages when training 7B models on 32 GPUs and 
    10K dataset initially, then the output length is increased linearly to 20K dataset.}
    \vspace*{-0.1in}
    \label{fig:eval:elastic}
\end{figure}

To show the effectiveness of the dynamic adjustment algorithm, we deploy \sysname-Async on 32 GPUs 
to train the 7B model and linearly scale the output length of the dataset to adjust the maximum 
output length from 10K to 20K. Figure~\ref{fig:eval:elastic} shows the delta time between the 
two stages in each iteration. As shown, after iteration 10 and 16, \sysname detects the imbalance and 
automatically adds one node with 8 GPUs to the \SGS stage to restore stage balance.

\subsection{Cross-Datacenter and Heterogeneity}
\label{sec:eval:heterogeneity}

\begin{figure}[t!]
    \centering
    \includegraphics[width=0.95\linewidth]{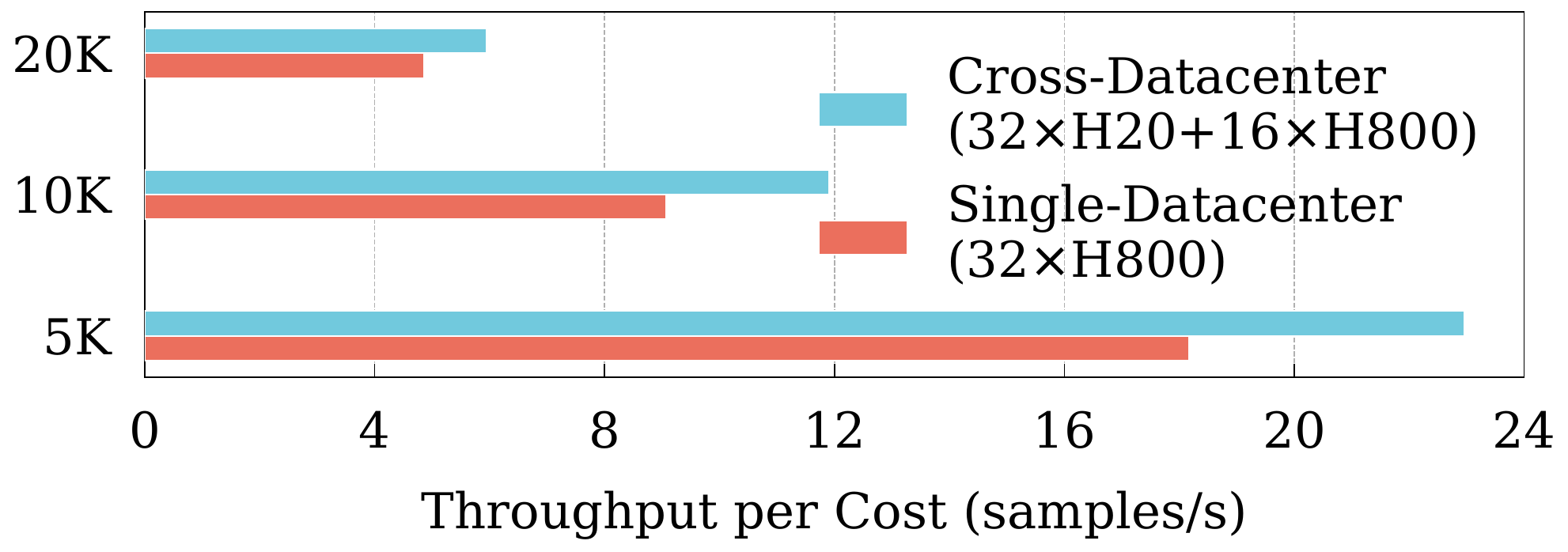}
    \vspace*{-0.15in}
    \caption{The throughput normalized by the hardware cost between cross- and single-datacenter
    deployment.}
    \vspace*{-0.1in}
    \label{fig:eval:hetero}
\end{figure}

As discussed in \S\ref{fig:background:motivation}, one promising potential of disaggregation 
lies in enabling each stage to utilize the most suitable hardware resources and supporting 
cross-datacenter training. To demonstrate this, we adopt the same settings as the end-to-end 
experiment with the 7B model, but move the \SGS of \sysname into a cloud-based H20 cluster with 32 GPUs.
\Trainer is still placed in the H800 cluster. 
We compare its performance against the original single datacenter setting. 
As shown in Figure~\ref{fig:eval:hetero}, with heterogeneous deployment, \sysname achieves 
a $1.23\times$--$1.31\times$ higher throughput normalized by hardware cost.
This improvement comes from the higher cost-efficiency of H20 for generation workloads. 
Additionally, the communication overhead introduced by cross-datacenter communication is small: 
each iteration only requires communication during weights updates, and even for a 72B model, 
the transmission overhead over a 80 Gbps dedicated link is less than 10 seconds—under 2\% 
of the total iteration time.

\subsection{Algorithmic Behavior of Asynchronous RL}
\label{sec:eval:async}

\begin{figure}[t!]
    \centering
    \includegraphics[width=0.95\linewidth]{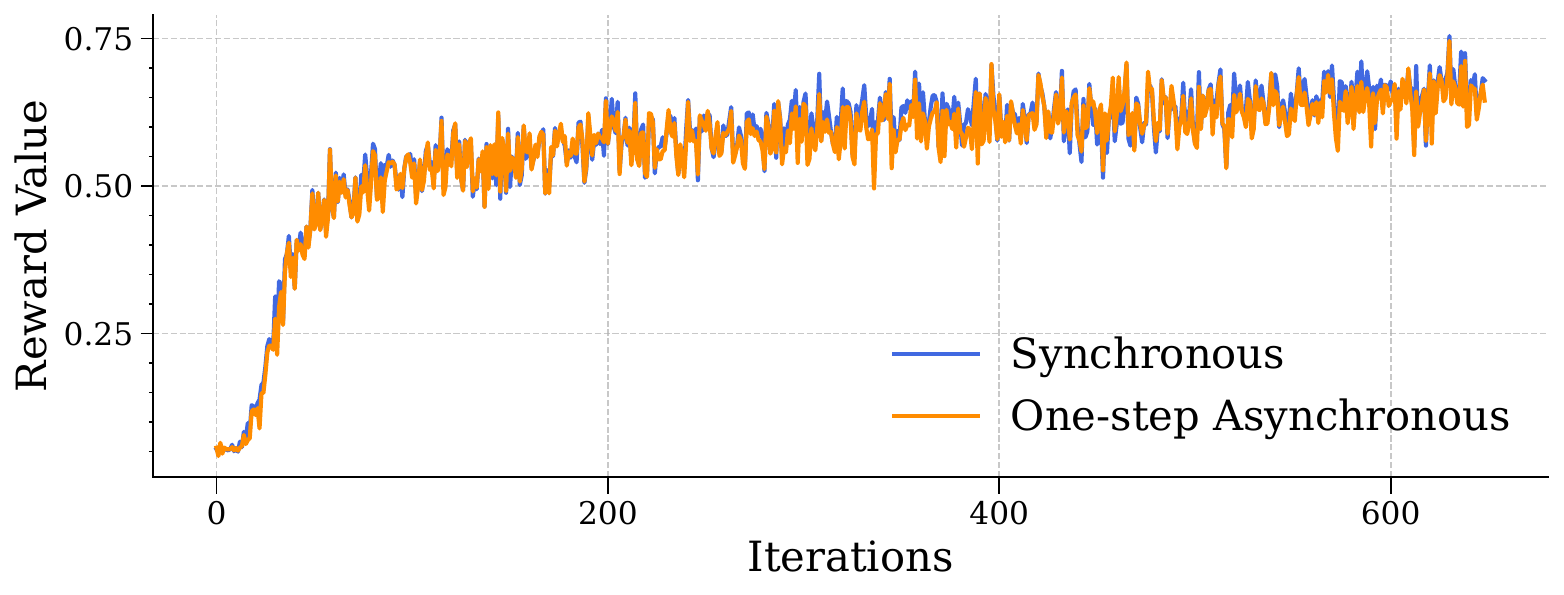}
    \vspace*{-0.2in}
    \caption{The reward curves between synchronous and one-step asynchronous PPO when training a 32B LLM.}
    \vspace*{-0.1in}
    \label{fig:eval:accuracy}
\end{figure}

The effectiveness of asynchronous RL for LLMs has been observed and validated by several 
prior works~\cite{noukhovitch2025asynchronousrlhffasterefficient, wang2025distrlasynchronousdistributedreinforcement, team2025kimi}.
To confirm this, we also conduct PPO-based RL training using Qwen2.5-32B~\cite{yang2024qwen2} as the base model on 
an internal dataset. keeping all other settings the same, Figure~\ref{fig:eval:accuracy} shows 
that the reward curves of the one-step asynchronous version closely match that of the synchronous version.
This demonstrates that it is possible to maximize training efficiency through algorithm-system 
co-design without compromising model performance and convergence. 

Of course, this case study only empirically verifies the feasibility of asynchronous training 
for specific LLM tasks; its generality and theoretical guarantees are beyond the scope of this paper.
Nevertheless, even if one has concerns about the potential model convergence problem introduced by asynchrony, 
you can use disaggregation with streaming to improve efficiency without changing the training semantics.

\section{Related Work}
\label{sec:related}

\paraf{RL training frameworks.} 
RL training is becoming increasingly important for LLMs to improve their performance and 
align their value with humans. To accelerate this process, various RL training frameworks 
have been proposed. One class of frameworks, such as NeMo~\cite{Harper_NeMo_a_toolkit} and 
OpenRLHF~\cite{hu2024openrlhf}, partitions the GPU cluster into multiple subsets to serve 
different stages of RL training. They are inefficient as only one stage can be executed 
simultaneously, leading to resource idleness. In contrast, verl~\cite{verl}, 
RLHFuse~\cite{zhong2024rlhfuse}, ReaL~\cite{mei2024realhf}, and PUZZLE~\cite{lei2024puzzle} 
colocate different stages on the same GPU pool to maximize resource utilization. 
ReaL~\cite{mei2024realhf} avoids under-utilization by parameter reallocation. 
RLHFuse~\cite{zhong2024rlhfuse} further proposes stage fusion to reduce the idleness at the 
sub-stage level. PUZZLE~\cite{lei2024puzzle} proposes lightweight context switching to reduce 
the switching overhead. However, they suffers from the resource coupling problem which is solved
by \sysname.

\parabf{LLM inference optimizations.}
Many optimizations have been proposed to accelerate the LLM inference. They are also applicable 
to the generation phase of RL training. ORCA~\cite{yu2022orca} proposes selective batching to batch 
requests with different lengths. vLLM~\cite{vllm} proposes PagedAttention to reduce memory 
fragmentation of various requests. FastServe~\cite{wu2023fast} uses preemptive scheduling to 
reduce the head-of-line blocking problem of long requests. Splitwise~\cite{patel2024splitwise} 
and DistServe~\cite{zhong2024distserve} split the prefill and decoding phases to avoid interference 
between them. Similar to \sysname, they also adopt the idea of resource disaggregation, but in the context of LLM inference. LoongServe~\cite{wu2024loongserve} proposes elastic sequence parallelism to serve 
different requests with different degrees of parallelism. They are orthogonal to \sysname and 
most of them are implemented in \sysname to improve the generation. 

\parabf{LLM training optimizations.}
LLM training has been extensively studied in the past few years.
Tensor parallelism~\cite{shoeybi2020megatronlm}, data parallelism, and pipeline 
parallelism~\cite{huang2019gpipe} are widely used to parallelize the LLM training in 
different dimensions. Alpa~\cite{alpa} proposes a unified framework to automatically
search for the optimal parallel strategies. CoDDL~\cite{coddl}, Pollux~\cite{pollux}, 
and ElasticFlow~\cite{gu2023elasticflow} elastically adjust the parallelism strategy 
to adapt to the workload. MegaScale~\cite{jiang2024megascale} summarizes various best 
practices for optimizing ultra-scale training. \sysname targets the RL training, 
where LLM training is just a single stage of the whole process.
\section{Conclusion}
\label{sec:conclusion}

In this work, we revisit the disaggregated architecture for RL training to highlight its promising 
advantages over the widely adopted colocation architecture: flexible resource allocation, 
support for heterogeneous hardware, and cross-datacenter scalability. To fully unlock the 
potential of disaggregation, we present \sysname, which addresses the pipeline bubbles and 
skewness-induced inefficiencies present in existing disaggregated RL frameworks. 
Experiments show that \sysname achieves up to a $2.66\times$ speedup compared to 
the current state-of-the-art RL framework. We hope this work encourages the community to 
revisit disaggregation and gain a deeper understanding of its effectiveness.

\bibliographystyle{style/ACM-Reference-Format}
\bibliography{paper}

\end{document}